\documentclass[11pt]{article}

\usepackage[preprint]{acl}

\usepackage{times}
\usepackage{latexsym}
\usepackage{pgfplots}
\usepackage{pgfplotstable}
\usepackage{booktabs}
\usepackage{verbatim}
\usepackage{float}
\pgfplotsset{compat=newest}
\usepackage{xparse}
\usepackage[most]{tcolorbox}

\usepackage[T1]{fontenc}

\usepackage[utf8]{inputenc}
\usepackage{comment}
\usepackage{xspace}
\newcommand{\methodname}{\textbf{LATTE}\xspace}

\usepackage{microtype}

\usepackage{inconsolata}

\usepackage{graphicx}
\usepackage{amsmath}
\usepackage{amssymb}

\title{LATTE: Learning Aligned Transactions and Textual Embeddings for Bank Clients}

\author{
  \textbf{Egor Fadeev\textsuperscript{1}},
  \textbf{Dzhambulat Mollaev\textsuperscript{1}},
  \textbf{Aleksei Shestov\textsuperscript{1}},
  \textbf{Omar Zoloev\textsuperscript{1, 3}},
  \textbf{Artem Sakhno\textsuperscript{1}},
  \textbf{Dmitry Korolev\textsuperscript{1}},\\
  \textbf{Ivan Kireev\textsuperscript{1}},
  \textbf{Andrey Savchenko\textsuperscript{1,2}},
  \textbf{Maksim Makarenko\textsuperscript{1}}\\
  \textsuperscript{1}Sber AI Lab, Moscow, Russia\\
 \textsuperscript{2}ISP RAS Research Center for Trusted Artificial Intelligence, Moscow, Russia\\
 \textsuperscript{3}NUST MISIS, Moscow, Russia 
  }

\begin{document}
\maketitle
\begin{abstract}
Learning clients embeddings from sequences of their historic communications is central to financial applications. While large language models (LLMs) offer general world knowledge, their direct use on long event sequences is computationally expensive and impractical in real-world pipelines. In this paper, we propose \methodname, a contrastive learning framework that aligns raw event embeddings with description-based semantic embeddings from frozen LLMs. Behavioral features  based on statistical user descriptions are summarized into short prompts, embedded by the LLM, and used as supervision via contrastive loss. The proposed approach significantly reduces inference cost and input size compared to the conventional processing of complete sequences by LLM. We experimentally show that our method outperforms state-of-the-art techniques for learning event sequence representations on real-world financial datasets while remaining deployable in latency-sensitive environments.
\end{abstract}

\section{Introduction}
Research in natural language processing (NLP) has traditionally focused on unstructured text~\cite{bagheri2023natural}. 
In contrast, many industrial applications of healthcare~\cite{wang2024twin}, education~\cite{Liu2023XES3G5M}, e-commerce~\cite{Dai2023, liu2025enhancing}, and, especially, finance~\cite{babaev2019rnn, luetto2025one}, generate  hundreds of streams of structured event (temporally ordered, high-dimensional, and often sparse tabular) data~\cite{osin2025ebeseasybenchmarkingevent}, such as transaction logs, payment histories, and customer interactions, which are sequential, high-dimensional, and sparse~\cite{zhang2023fatatrans, osin2025ebeseasybenchmarkingevent}. These data underpin a broad spectrum of business-critical tasks, including churn prediction, risk assessment, credit scoring, and personalized targeting~\cite{mollaev2025multimodalbankingdatasetunderstanding}. Transaction sequences differ from text in three key ways: they are much longer (thousands of events in open datasets, millions in proprietary banking logs)~\cite{mollaev2025multimodalbankingdatasetunderstanding}, each event includes multiple categorical and numerical attributes~\cite{zhang2023fatatrans}, and the main tasks are classification or regression rather than broad semantic benchmarks~\cite{muennighoff2023mteb}.

\begin{figure}
    \centering
    \includegraphics[width=\linewidth]{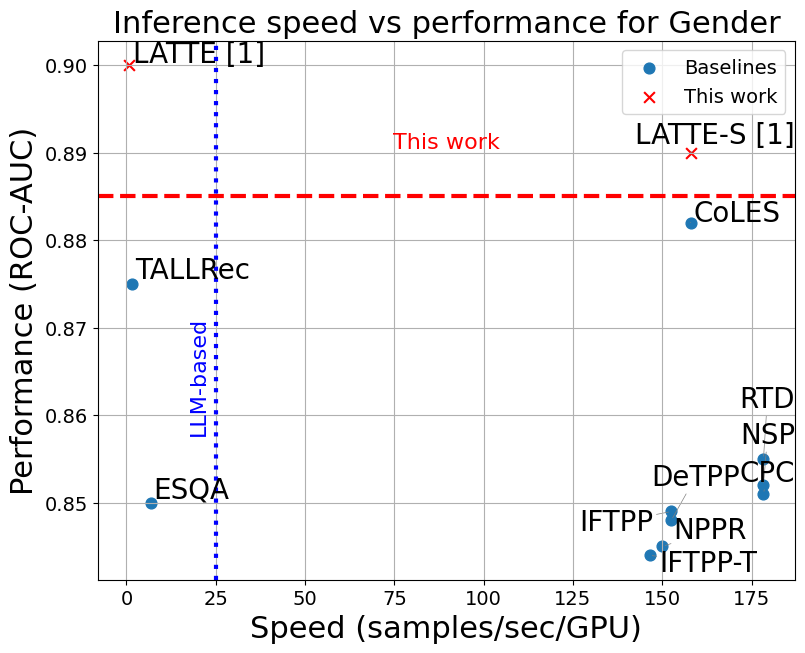}
    \caption{
        Figure of Merit comparing ROC-AUC performance and inference speed (samples/sec/GPU) on the Gender prediction task.
        Compared methods include \methodname[\ref{eq:softmax}], \methodname-S[\ref{eq:softmax}], CoLES~\cite{babaev2022coles} RTD~\cite{clarkelectra}, CPC~\cite{oord2018representation}, NSP~\cite{devlin2019bert}, NPPR~\citep{skalski2023towards}, DeTPP~\cite{karpukhin2024detpp}, IFTPP~\cite{shchurintensity}, IFTPP-T~\cite{shchurintensity}, ESQA~\cite{abdullaeva2024esqa}, and TALLRec~\cite{bao2023tallrec}.
    }
    \label{fig:fom}
\end{figure}

Recent works on applying Large Language Models (LLMs) for structured data~\cite{shi2023language,yu2025eventformer} highlight that progress in this domain depends on methods adapted to the unique statistical and causal structure of event data, rather than direct transfer of techniques from NLP. Moreover, a direct application of LLMs to serialized sequential tabular data incurs substantial computational overhead due to the large token counts per user. For example, typical banking transaction datasets often contain hundreds of records per user, each serialized into dozens of tokens, exceeding practical context window limits and increasing inference and training times~\cite{shestov2025llm4es}. Figure~\ref{fig:fom} shows that LLM-based models such as TALLRec~\cite{bao2023tallrec} and ESQA~\cite{abdullaeva2024esqa} do not exceed an inference speed of 10 users per second, which severely limits their applicability in banking production environments.

\begin{figure*}[t]
    \centering
    \includegraphics[width=\textwidth]{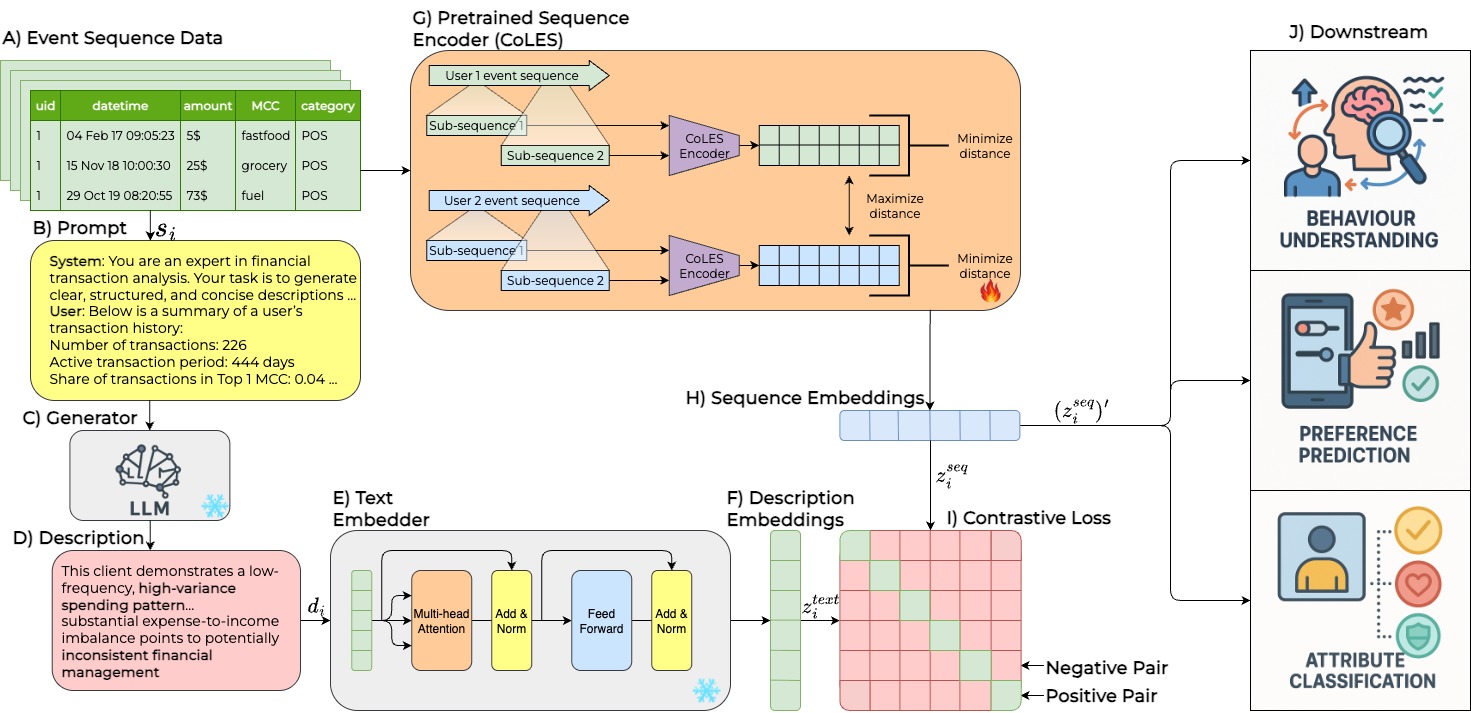}
    \caption{Overview of the proposed \methodname pipeline. 
    (a) Event sequences serve as the input data source. 
    (b) A summary prompt is crafted to query the event sequence. 
    (c) An LLM generator produces a natural language description based on the prompt. 
    (d) The resulting textual description captures salient features of the event sequence. 
    (e) A text embedder converts the description into a vector representation. 
    (f) This description embedding encodes the generated text. 
    (g) In parallel, a sequence encoder embeds the original event sequence. 
    (h) The resulting sequence embedding captures structural and temporal patterns. 
    (i) A contrastive alignment module trains the model to align textual and sequence embeddings in a shared representation space. 
    (j) The aligned embeddings can be used for various downstream tasks such as classification, retrieval, or prediction.}
    \label{fig:pipeline}
\end{figure*}

To address the limitations of existing techniques, we propose \methodname, a scalable framework for Learning Aligned Transactions and Textual Embeddings. Instead of feeding entire sequences into LLMs, we extract compact client-level statistics and use an instruction-tuned LLM to generate natural language summaries. These summaries serve as weak labels, aligned with pretrained sequence embeddings from a lightweight encoder via contrastive learning. At inference, \methodname supports two modes: a standalone encoder (\methodname-S) that retains LLM-level semantics without added overhead, and a combined encoder (\methodname) that fuses textual representations enriched with statistical features and structural embeddings.

To evaluate trade-offs between performance and efficiency, we introduce a Figure of Merit (FOM) comparing ROC-AUC and inference speed on the standard banking dataset of gender prediction task~\citep{sberbank2021gender} (Figure~\ref{fig:fom}). We evaluate \methodname in two variants, combining two embedding strategies: structural-only (\methodname-S) and concatenated with textual features (\methodname). In terms of performance, all versions of \methodname outperform the typical baseline financial methods. The structural-only variant (\methodname-S) achieves a ROC-AUC of 0.891 on the Gender task with an inference speed of 162 samples/sec/GPU, surpassing LLM-based models such as TALLRec and ESQA, being over $14\times$ faster. Across all types of textual encoders, the combined variant (\methodname) consistently achieves the highest overall ROC-AUC.

\section{Related Works}
Learning representations from structured event sequences~\cite{Udovichenko_2024, yeshchenko2022surveyapproacheseventsequence, system_call_sequences, Rare_Events, guo2020surveyvisualanalysisevent} remains a core challenge in industrial applications. Despite abundant customer interaction data, high-quality labels for event sequences in most typical downstream tasks (campaigning, churn prediction, etc.) remain limited~\cite{mollaev2025multimodalbankingdatasetunderstanding}. This shortage of timely supervision hinders the scalability of supervised learning in production settings. It highlights the need for self-supervised approaches to derive robust and semantically rich representations directly from raw behavioral sequences~\cite{gui2024survey}. Prevailing self-supervised approaches for modeling event sequences adopt such objectives as contrastive learning~\cite{babaev2022coles}, next-event prediction~\cite{skalski2023towards}, and latent sequence modeling techniques, e.g., Contrastive Predictive Coding (CPC)~\cite{oord2018representation}, aiming to capture temporal dependencies and user intent without relying on manual supervision.

Appearance of LLMs offer new opportunities to enhance representation learning from event sequences. Trained on diverse and large-scale corpora, LLMs encode elements often implicit or absent in structured event datasets, e.g., rich semantic priors about behavioral patterns, temporal dynamics, and domain knowledge. Leveraging this external knowledge can significantly improve the quality of user representations, particularly in financial applications~\cite{ruan2024languagemodelingtabulardata}. Motivated by this potential, recent studies have explored adaptations of LLMs to structured data. For example, TALLRec~\cite{bao2023tallrec}, LLM-TRSR~\cite{zheng2024harnessing}, and HKFR~\cite{yin2023heterogeneous} transfer rich text understanding abilities of LLMs to recommender systems; TabLLM~\cite{hegselmann2023tabllm} targets tabular classification tasks; TEST~\cite{suntest} and Time-LLM~\cite{jintime} address time series; while ESQA~\cite{abdullaeva2024esqa} applies LLMs to event-sequence question answering.

Existing methods to mitigate this issue fall into two main categories. The first reduces context length by summarizing user histories with general-purpose LLMs~\cite{yin2023heterogeneous,zheng2024harnessing}, which risks losing domain-specific information critical for accurate modeling. The second class of methods attempts to bypass context length constraints by encoding sequences in non-textual formats~\cite{suntest,jintime}. However, these representations often discard the semantic content present in item descriptions (e.g., transaction categories or merchant details). Moreover, as user histories grow longer, these models either incur increased computational cost or suffer performance degradation due to limited model capacity.

Recent advances in event sequence modeling reinforce the distinction between unstructured texts from NLP and structured data. Work on spatio-temporal clustering shows that standard neural point process models fail to capture hierarchical spatial structures and multi-type dependencies, requiring new architectures tailored to these properties~\cite{yu2025eventformer}. Other studies demonstrate that even transformer-based approaches underperform when causal relations between event types are ignored, motivating causality-aware attention mechanisms~\cite{shou2023pairwise}. Further results indicate that while Large Language Models (LLMs) can aid event prediction through abductive reasoning, they are effective only when combined with specialized sequence models~\cite{shi2023language}. Together, these findings highlight that progress in this domain depends on methods adapted to the unique statistical and causal structure of event data, rather than direct transfer of techniques from NLP.

\section{Proposed Approach}

We aim to improve the quality of representations learned from transactional event sequences by introducing an auxiliary textual modality that verbalizes statistical properties of user behavior. To this end, we propose a three-stage pipeline, \methodname, illustrated in Figure~\ref{fig:pipeline}, which maps raw transaction sequences into rich embeddings suitable for downstream tasks.

Let $T_i = {x_1, x_2, \dots, x_n}$ denote the transaction sequence for client $i$, where each $x_j$ contains timestamped categorical and numerical attributes (e.g., amount, merchant category). As shown in Fig.~\ref{fig:pipeline}a, we first compute a vector of summary features $s_i$ that aggregates behavioral patterns over the sequence: frequency of activity, merchant diversity, transaction types, temporal coverage, and income-expense structure. Behavioral features are then transformed into meaningful textual descriptions rather than raw indices, allowing these summaries to be further enriched with the semantic knowledge of the LLM. The prompt template and a sample generated description are presented in Appendix~\ref{sec:appendix_generation}. This profile is then serialized into a textual prompt (Fig.~\ref{fig:pipeline}b), which is passed to a frozen instruction-tuned LLM (Fig.~\ref{fig:pipeline}c) to generate a natural language description $d_i$ of the client’s behavior (Fig.~\ref{fig:pipeline}d). The prompt includes a system message and a structured representation of $s_i$ designed to elicit coherent and interpretable responses.

Simultaneously, the raw sequence $T_i$ is processed by a GRU-based sequence encoder (Fig.~\ref{fig:pipeline}G) trained under a self-supervised CoLES objective~\citep{babaev2022coles}, where each training example consists of two overlapping subsequences (positives) and contrastive negatives from other clients. This produces the sequence embedding $z_i^{\text{seq}}$ optimized to capture client-specific behavioral dynamics. In parallel, the description $d_i$ is passed through a frozen multilingual sentence encoder (Fig.~\ref{fig:pipeline}f) with mean pooling, producing the text embedding $z_i^{\text{text}}$ enriched with raw statistical features derived from the original sequence. 

The embeddings $z_i^{\text{seq}}$ and $z_i^{\text{text}}$ are then aligned using one of three cross-modal contrastive losses (Fig.~\ref{fig:pipeline}i), while keeping the text encoder fixed. The updated transaction embeddings $\left(z_i^{\text{seq}}\right)^{'}$ (Fig.~\ref{fig:pipeline}h), aligned with the textual representations, are evaluated on downstream tasks such as churn prediction (Fig.~\ref{fig:pipeline}j). Appendix \ref{sec:appendixset} provides additional details regarding the pipeline. 

To perform cross-modal alignment, we introduce two different contrastive heads: Symmetric Softmax and Orthogonal Regularized. Each was inspired by prior work on multimodal learning~\citep{radford2021learning, jiang2023understanding}. These heads differ in their alignment geometry and regularization, and we refer to them as \methodname[\ref{eq:softmax}], and \methodname[\ref{eq:reg}], respectively. Both heads use frozen text encoders and update only the transaction encoder. Downstream tasks are evaluated using the resulting $\left(z_i^{\text{seq}}\right)^{'}$, while the textual modality provides semantically grounded alignment by enriching the training signal with contextual knowledge of categorical attributes that the sequence encoder alone cannot interpret.\\
\textbf{\methodname[\ref{eq:softmax}]: Symmetric Softmax Contrastive Head} promotes bidirectional alignment between modalities using a symmetric InfoNCE-style loss:
\begin{equation}
\label{eq:softmax}
\mathcal{L}_{\text{softmax}} = \frac{1}{2}(\mathcal{L}_{\text{seq} \rightarrow \text{text}} + \mathcal{L}_{\text{text} \rightarrow \text{seq}}),
\end{equation}
where

\begin{align*}
\mathcal{L}_{\text{seq} \rightarrow \text{text}} &= -\frac{1}{N} \sum_{i=1}^N \log \frac{\exp(\langle z_i^{\text{seq}}, z_i^{\text{text}} \rangle / \tau)}{\sum_{j=1}^N \exp(\langle z_i^{\text{seq}}, z_j^{\text{text}} \rangle / \tau)}. \
\end{align*}
The second term, $\mathcal{L}_{\text{text} \rightarrow \text{seq}}$, is defined similarly with the roles of sequence and text reversed. Here, $\langle \cdot, \cdot \rangle$ denotes the cosine similarity between the L2-normalized embeddings of the sequence ($z_i^{\text{seq}}$) and the corresponding text ($z_i^{\text{text}}$), and $\tau$ is a temperature hyperparameter controlling the sharpness of the similarity distribution.\\
\textbf{\methodname[\ref{eq:reg}]: Orthogonal Regularized Contrastive Head} augments the softmax-based loss with a geometric regularization term that disentangles modality-specific and shared information. To achieve this, we introduce an auxiliary projection head that maps each transaction embedding $z_i^{\text{seq}}$ to a representation composed of two parts: $Z^{\text{shared}}$, which captures components aligned with textual information, and $Z^{\text{spec}}$, which preserves information specific to the transaction modality: 

\begin{equation}
\label{eq:reg}
\mathcal{L}_{\text{reg}} = \mathcal{L}_{\text{softmax}} + \lambda_{\text{ortho}} \cdot \mathcal{L}_{\text{ortho}},
\end{equation}
where $\mathcal{L}_{\text{ortho}} = \left\lVert (Z^{\text{shared}})^\top Z^{\text{spec}} \right\rVert_F^2$.

This separation promotes disentangled features by penalizing correlation between shared and specific subspaces, where $\lambda_{\text{ortho}}$ controls the strength of this regularization and thus the emphasis on preserving modality-specific information. 

\section{Experimental Setup}

This section provides the core details of our experimental setup, including validation strategy, datasets, and baseline methods. Additional implementation details are provided in Appendix~\ref{sec:appendixset}.

\subsection{Data} We evaluate our method on three real-world datasets containing anonymized credit card transaction sequences from major financial institutions. Each dataset comprises client-level sequences with numerical and categorical attributes (e.g., amount, merchant category, transaction type), and includes an unlabeled subset used exclusively for representation learning.
\emph{Churn}~\citep{rosbank2021churn} includes approximately 10K Rosbank clients labeled by future inactivity. \emph{Gender}~\citep{sberbank2021gender} and \emph{Age Group}~\citep{sberbank2021age}, provided by Sberbank, contain 15K and 50K clients respectively, annotated with demographic labels.

\subsection{Validation Strategy}

Each dataset is split into disjoint training and test partitions by reserving 10\% of the labeled clients for evaluation. The remaining 90\% of labeled users, together with all available unlabeled users, are used for training the embedding models. To assess the quality of learned representations, we adopt a 5-fold cross-validation procedure. Specifically, the labeled portion of the training data is divided into five equal-sized folds. For each fold $v$, we: (1) train a LightGBM~\citep{ke2017lightgbm} classifier on embeddings from the remaining four folds, and (2) evaluate it on the held-out test fold, computing a downstream performance metric $M_v$. For binary classification tasks (churn, gender), we report ROC-AUC; for multiclass age prediction, we report classification accuracy. The final performance is summarized as $\mu_M \pm \sigma_M$, where $\mu_M$ is the mean and $\sigma_M$ is the standard deviation over all five folds.
\subsection{Baselines}

We compare \methodname against a diverse set of baselines spanning five methodological families:\\
\textbf{Event sequence models.} CPC~\citep{oord2018representation} learns to predict future representations from past context via a contrastive loss. CoLES~\citep{babaev2022coles} improves temporal consistency by aligning overlapping subsequences using InfoNCE. NPPR~\citep{skalski2023towards} employs autoregressive training with dual objectives: predicting the next and reconstructing the previous event from masked sequences. All models operate on raw event sequences without external modalities.\\
\textbf{Temporal point process models.} DeTPP~\citep{karpukhin2024detpp} models event timing and types using parametric point processes. IFTPP and IFTPP-T~\citep{shchurintensity} use Transformer and GRU backbones with combined MAE and classification losses.\\
\textbf{Natural language processing objectives.} RTD~\citep{clarkelectra} randomly replaces 15\% of event tokens and predicts the original token; NSP~\citep{devlin2019bert} extends BERT's next sentence prediction to sequences of events.\\
\textbf{LLM-based approaches.} 
We adapt TALLRec~\cite{bao2023tallrec} and HKFR~\cite{yin2023heterogeneous} from recommender systems for user embedding extraction by fine-tuning LLMs on serialized user-item histories using next-token prediction. Embeddings are derived via mean pooling over the final layer, following recent best practices~\cite{behnamghader2024llmvec,muennighoff2024generative}. Additionally, we replace older backbones (e.g., LLaMA 7B~\cite{touvron2023llama}, ChatGLM-6B~\cite{du2022glm}) with LLaMA 3.2 3B~\cite{touvron2024llama3} to leverage architectural improvements and efficiency.\\
\textbf{Tabular feature aggregation.} As a non-sequential baseline, \textit{agg} aggregates transaction features using summary statistics such as mean, standard deviation, min-max and grouped frequency statistics (for categorical features).

\section{Experimental Results}
\subsection{Main Results}

\begin{table}[hbtp]
\centering
\resizebox{\columnwidth}{!}{
\begin{tabular}{lccc}
\hline
\textbf{Model} & \textbf{Churn (AUC)} & \textbf{Age Group (Acc)} & \textbf{Gender (AUC)} \\
\hline
agg & 0.827 ± 0.010 & 0.629 ± 0.002 & 0.877 ± 0.004 \\
CPC & 0.792 ± 0.015 & 0.602 ± 0.004 & 0.851 ± 0.006 \\
RTD & 0.771 ± 0.016 & 0.631 ± 0.006 & 0.855 ± 0.008 \\
CoLES & 0.841 ± 0.005 & 0.644 ± 0.005 & 0.882 ± 0.004 \\
NSP & 0.828 ± 0.012 & 0.621 ± 0.005 & 0.852 ± 0.011 \\
NPPR & 0.845 ± 0.003 & 0.642 ± 0.001 & - \\
DeTPP & 0.823 ± 0.002 & 0.632 ± 0.004 & - \\
IFTPP & 0.828 ± 0.004 & 0.632 ± 0.003 & 0.863 ± 0.003 \\
IFTPP-T & 0.814 ± 0.004 & 0.620 ± 0.002 & 0.852 ± 0.005 \\
TALLRec & 0.839 $\pm$ 0.003 & 0.659 ± 0.004 & 0.875 $\pm$ 0.004 \\
HKFR & 0.823 $\pm$ 0.006 & - & - \\
\hline
\methodname[\ref{eq:softmax}] & \underline{0.869 ± 0.004} & \textbf{0.665 ± 0.005} & \textbf{0.900 ± 0.005} \\
\methodname[\ref{eq:reg}] & \textbf{0.872 ± 0.004} & \underline{0.663 ± 0.003} & \underline{0.898 ± 0.006} \\
\hline
\end{tabular}
}
\caption{Performance of client embeddings on downstream tasks. \textbf{Bold} indicates best result; \underline{underline} indicates second-best}
\label{tab:main_results}
\end{table}

Table~\ref{tab:main_results} presents the performance of client embeddings on three downstream tasks: churn prediction (AUC), age group classification (accuracy), and gender prediction (AUC). While traditional self-supervised objectives such as CPC, RTD, and NSP lag behind stronger baselines like CoLES and TALLRec, several variants of our proposed LATTE framework demonstrate clear gains across tasks. \methodname[\ref{eq:reg}] achieves the best result on churn prediction (0.872 AUC), while \methodname[\ref{eq:softmax}] leads on both age group classification (0.665 accuracy) and gender prediction (0.900 AUC). These improvements indicate that incorporating statistic-based textual supervision leads to consistently stronger representations of transactional behavior.

\subsection{Runtime Analysis}

\begin{figure}[h]
    \centering
    \includegraphics[width=\linewidth]{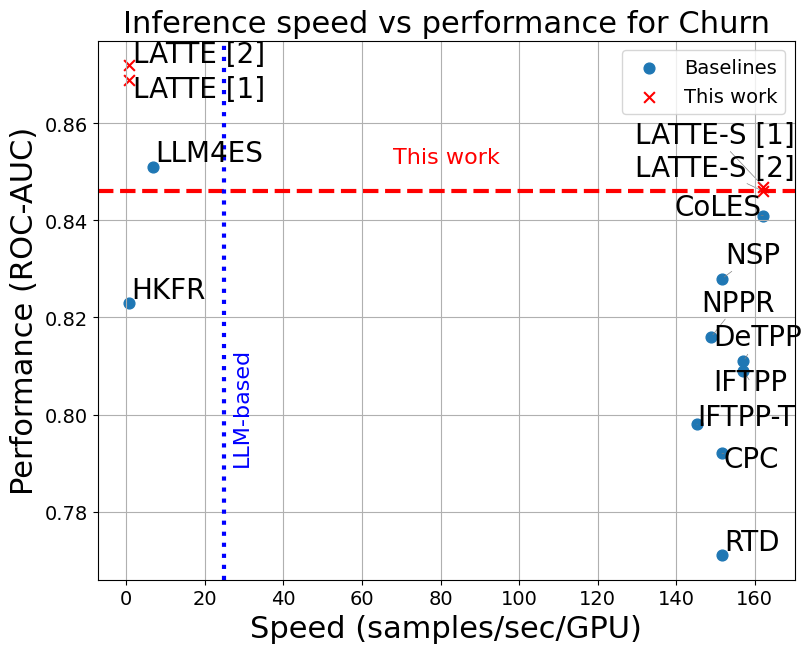} 
    \caption{
        Figure of Merit comparing model performance in ROC-AUC and inference speed in samples per second per GPU on Churn dataset. 
    }
    \label{fig:churn}
\end{figure}

In banking applications, inference speed is critical due to the need to process millions of multi-source user records in a hardware environment with limited access to GPUs. In this section, we investigate an important trade-off between inference resource utilization (compute time, memory utilization) and the performance of the proposed methods compared to baseline architectures. 

Figure~\ref{fig:churn} showcases the performance on the Churn dataset. Generative models (\methodname, TALLRec) achieve the highest ROC-AUC scores, exceeding 0.868, but at the cost of extremely low inference speeds—processing only a few samples per second—and high memory consumption, with over 3 billion parameters. In contrast, lightweight \methodname-S variants attain slightly lower ROC-AUC values while operating over 160 samples per second and a compact size of just a few million parameters These models match the efficiency of contrastive baselines, offering a favorable trade-off between accuracy, speed, and memory footprint.

\subsection{Contrastive Fine-tuning Alignment}

\begin{table}[h]
\centering
\resizebox{\columnwidth}{!}{
\begin{tabular}{lccc}
\hline
\textbf{Method} & \textbf{Churn (AUC)} & \textbf{Age Group (Acc)} & \textbf{Gender (AUC)} \\
\hline
Descriptions & 0.772 ± 0.011 & 0.432 ± 0.007 & 0.644 ± 0.009 \\
$z^{\text{seq}}$ + $z^{\text{text}}$ & 0.863 ± 0.007 & 0.650 ± 0.004 & 0.890 ± 0.003 \\
\methodname-S[\ref{eq:softmax}] & 0.847 ± 0.004 & 0.657 ± 0.003 & 0.891 ± 0.004 \\
\methodname[\ref{eq:softmax}] & \underline{0.869 ± 0.004} & \textbf{0.665 ± 0.005} & \textbf{0.900 ± 0.005} \\
\methodname-S[\ref{eq:reg}] & 0.846 ± 0.005 & 0.655 ± 0.004 & 0.888 ± 0.003 \\
\methodname[\ref{eq:reg}] & \textbf{0.872 ± 0.004} & \underline{0.663 ± 0.003} & \underline{0.898 ± 0.006} \\
\hline
\end{tabular}
}
\caption{Ablation: Impact of contrastive fine-tuning and modality concatenation on downstream task.}
\label{tab:ablation_finetune_concat}
\end{table}

Table~\ref{tab:ablation_finetune_concat} presents an ablation study on the effect of contrastive fine-tuning and modality concatenation. Incorporating LLM-generated behavioral descriptions markedly improves downstream performance compared to CoLES, particularly for churn and age prediction tasks. The contrastive alignment step remains crucial: all \methodname variants outperform the non-aligned baseline. Among the evaluated heads, \methodname[\ref{eq:reg}] achieves the highest AUC on churn (0.872), while \methodname[\ref{eq:softmax}] attains the best accuracy on age (0.665) and and gender (0.900). Nevertheless, unaligned concatenation remains a competitive baseline ($z^{\text{seq}}$ + $z^{\text{text}}$ ), indicating that the statistical–semantic descriptions alone already provide a strong inductive bias.

\subsection{Evaluation of the quality of LLM summarization}

\begin{table}[h]
\centering
\resizebox{\columnwidth}{!}{
\begin{tabular}{lccc}
\hline
\textbf{Feature} & \textbf{Churn} & \textbf{Gender} & \textbf{Age} \\
\hline
\textbf{mcc\_0 Usage \%} & 35.59 & 24.61 & 37.93 \\
\textbf{mcc\_0 Acc \%} & 100 & 100 & 100 \\
\textbf{mcc\_1 Usage \%} & 38.98 & 26.61 & 39.66 \\
\textbf{mcc\_1 Acc \%} & 100 & 100 & 100 \\
\textbf{trx\_period Usage \%} & 100 & 100 & 100 \\
\textbf{trx\_period Acc \%} & 98.31 & 99.02 & 99.14 \\
\textbf{trx\_days\_share Usage \%} & 93.22 & 95.41 & 93.97 \\
\textbf{trx\_days\_share Acc \%} & 98.18 & 92.38 & 99.07 \\
\hline
\end{tabular}
}
\caption{Four key LLM statistics usage (\%) and accuracy (\%) across tasks.}
\label{tab:feature_usage_accuracy_transposed}
\end{table}

In this section, we study whether the textual descriptions generated by the LLM faithfully capture the underlying statistics that were used to construct the prompts. We asked an independent LLM critic (Llama 3.1 8B) to extract key statistics (e.g., dominant merchant categories, transaction period length, share of active days) from the subsample of generated descriptions. We then applied a rule-based matching procedure to compare these extracted factors against the ground-truth statistics.

Table \ref{tab:feature_usage_accuracy_transposed} reports both the usage rate (how frequently a given statistic was mentioned in the LLM description) and the accuracy rate (the percentage of mentions that correctly reflect the underlying value) across three downstream tasks. We use 200 random samples per dataset. The results show that core statistics such as transaction\_period and transaction\_days\_share are not only used very frequently (over 90\% of cases), but also described with a high accuracy (above 92\%). As expected, categorical statistics such as merchant-category (mcc\_0, mcc\_1) are consistently mentioned with 100\% correctness when they appear.

\begin{figure*}[ht]
    \centering
    \includegraphics[width=\linewidth]{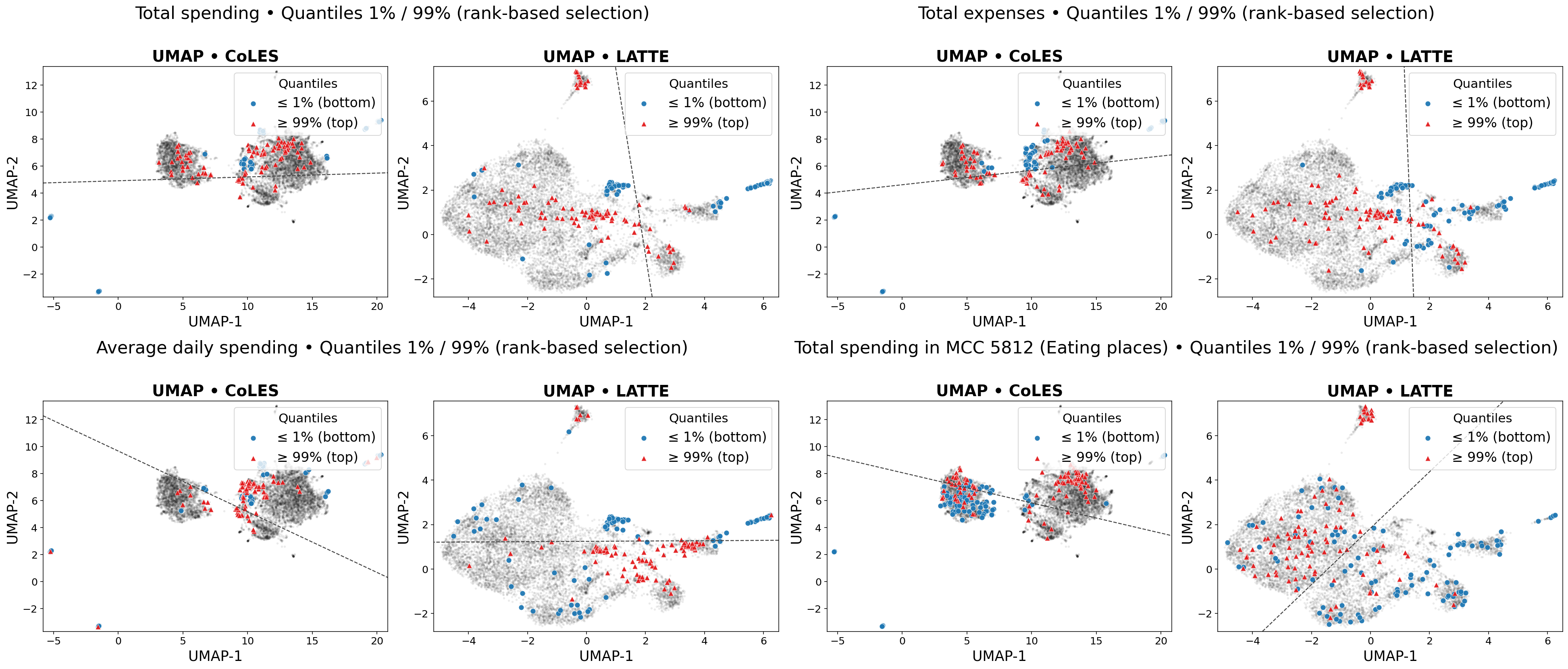}
    \caption{UMAP visualizations of CoLES (left) and LATTE (right) embeddings colored by quantiles of different behavioral statistics. LATTE embeddings exhibit slightly better linear separation across some statistics such as (a) Total spending, (b) Total expense, (c) Average daily spending, and (d) Total spending in MCC (Eating places).}
    \label{fig:umap_statistics}
\end{figure*}

\section{Analysis of Behavioral Embedding Structure}

\begin{table}[h]
\centering
\resizebox{\columnwidth}{!}{
\begin{tabular}{lcc}
\hline
\textbf{Metric} & \textbf{CoLES} & \textbf{LATTE} \\ \hline
Total spending & 0.627 & \textbf{0.693} \\
Total expense & 0.593 & \textbf{0.730} \\
Average daily spending & 0.519 & \textbf{0.689} \\
Total spending in MCC (Eating places) & 0.705 & \textbf{0.775} \\
First transaction day & \textbf{0.652} & 0.622 \\
Std of daily spending & \textbf{0.858} & 0.789 \\
\hline
\end{tabular}
}
\caption{Separability of user groups in the lowest 1\% and highest 99\% quantiles of behavioral statistics using logistic regression. LATTE embeddings show stronger separation across most statistics compared to CoLES.}
\label{tab:churn_coles_latte_transposed}
\end{table}

In this section, we investigate how behavioral statistics are encoded within the embedding spaces of CoLES and LATTE (Figure~\ref{fig:umap_statistics}). For each statistic, we highlight clients belonging to the lowest 1\% (blue) and highest 99\% (red) quantiles, and visualize the resulting structure using a UMAP projection with an overlaid logistic regression decision boundary. Compared to CoLES, LATTE embeddings exhibit more distinct geometric separation across representative behavioral dimensions-such as total spending, total expenses, average daily spending, and spending within the Eating places MCC category—indicating a stronger alignment between embedding geometry and behavioral variance.

To complement these qualitative observations, we conduct a quantitative assessment of separability for users with extreme behavioral profiles. As reported in Table~\ref{tab:churn_coles_latte_transposed}, LATTE consistently achieves higher separability scores than CoLES, particularly for total expenses, average spending, and spending in top MCC categories, while performance remains comparable for total sum and weekday spending. 

\begin{figure}[h!]
    \centering
    \includegraphics[width=\linewidth]{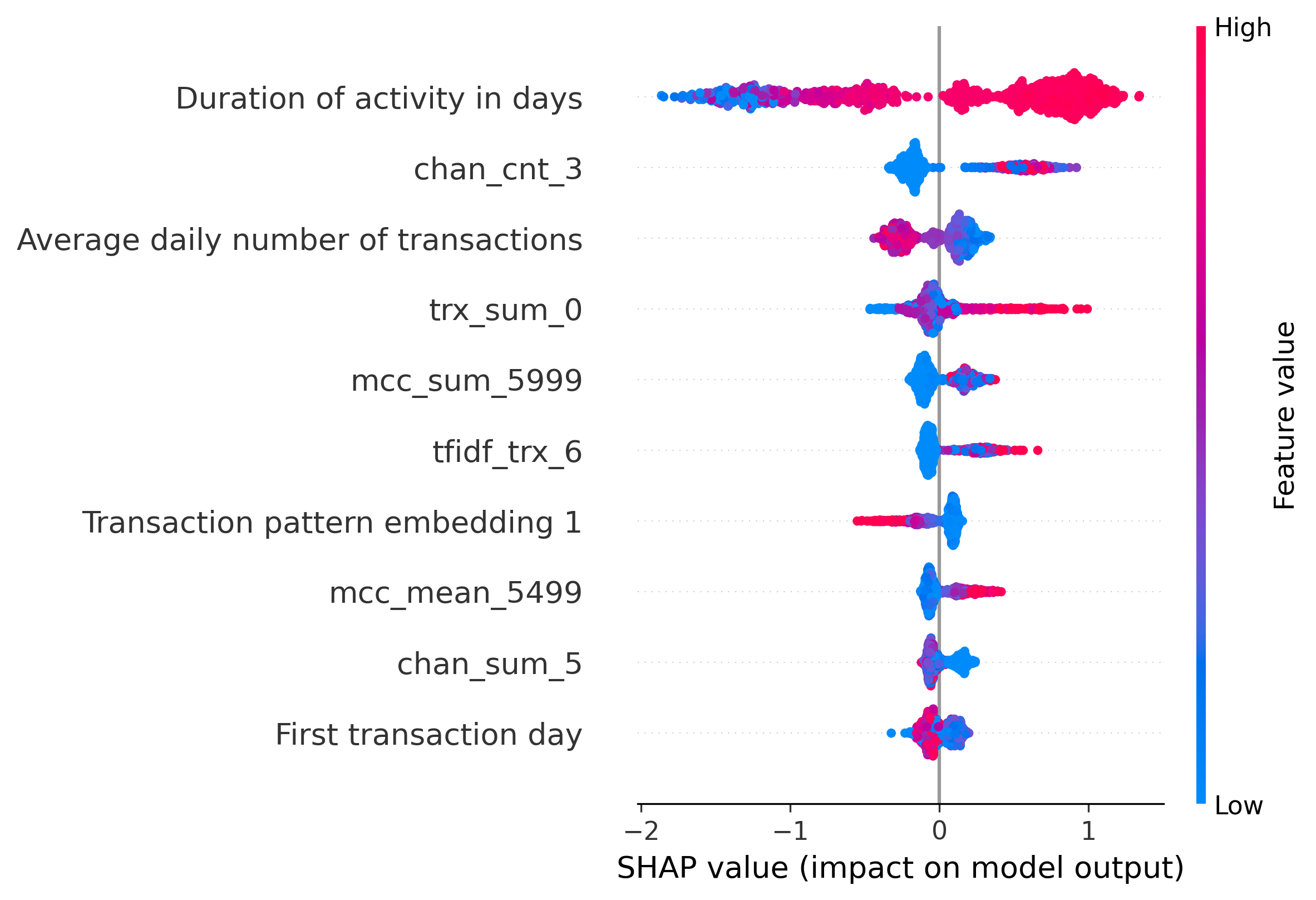}
    \caption{SHAP summary plot showing the most influential features contributing to model predictions.}
    \label{fig:shap_statistics}
\end{figure}

Finally, the SHAP summary plot (Figure~\ref{fig:shap_statistics}) provides further insight into feature importance within the predictive model. The most influential predictors include duration of activity, channel count, and average daily number of transactions, which show strong positive SHAP values for clients with higher feature magnitudes. Additionally, MCC-specific spending features and transaction-pattern embeddings make substantial contributions, implying that both aggregated behavioral indicators and categorical spending patterns play a crucial role in model predictions.

\section{Conclusion}
We presented a novel method (Fig.~\ref{fig:pipeline}) for contrastive representation learning from event sequences that leverages synthetically generated textual descriptions as a complementary modality. By aligning structured transaction data with natural language summaries produced by a frozen instruction-tuned LLM, the proposed approach introduces textual priors into the embedding space without requiring labeled data or LLM fine-tuning. Our \methodname achieves state-of-the-art results across three key open-source banking tasks, with relative improvements of 6.1\% in gender prediction, around 1.0\% in age group classification, and 3.7\% in churn prediction compared to the baseline. The proposed \methodname-S is resource-efficient (a few million parameters, up to 200 samples/sec speed), which is essential for industrial applications where behavioral logs are abundant but supervision is limited. 

A particularly promising future direction is to explore richer forms of text-to-sequence alignment, where natural language summaries are coupled with the underlying event dynamics. This approach could yield to inherently interpretable embeddings.

\section*{Limitations}

While our approach achieves strong empirical performance, it remains constrained by its reliance on a fixed set of pre-computed statistical features that condition the LLM-generated textual descriptions. This dependency limits adaptability when key behavioral patterns are not adequately captured by the chosen statistics. Furthermore, the method assumes that the generated descriptions faithfully reflect the underlying sequence dynamics, making performance sensitive to prompt design and the generalization capacity of the frozen LLM. Because the text encoder is not updated during training, alignment fidelity may further degrade under distributional shifts. Moreover, although the framework requires neither labels nor fine-tuning, it introduces moderate training overhead compared to lightweight contrastive objectives due to large-scale LLM-based generation.

Finally, in this paper, only financial transactions are used. The proposed sequence-to-text alignment framework can be extended to a wide range of domains that generate structured event logs-data types where LLMs often struggle due to sparsity, heterogeneity, and long temporal dependencies. Examples include healthcare~\cite{wang2024twin}, education~\cite{Liu2023XES3G5M}, e-commerce~\cite{Dai2023, liu2025enhancing}. In these settings, \methodname could leverage LLM-generated descriptions to inject semantic priors into structural representations.

\section*{Acknowledgments}
The work of A. Savchenko was supported by a grant, provided by the Ministry of Economic Development of the Russian Federation in accordance with the subsidy agreement (agreement identifier 000000C313925P4G0002) and the agreement with the Ivannikov Institute for System Programming of the Russian Academy of Sciences dated June 20, 2025 No. 139-15-2025-011.

\bibliography{custom}

\begin{thebibliography}{51}
\providecommand{\natexlab}[1]{#1}

\bibitem[{Abdullaeva et~al.(2024)Abdullaeva, Filatov, Orlov, Karpukhin, Vasilev, Dimitrov, Kuznetsov, Kireev, and Savchenko}]{abdullaeva2024esqa}
Irina Abdullaeva, Andrei Filatov, Mikhail Orlov, Ivan Karpukhin, Viacheslav Vasilev, Denis Dimitrov, Andrey Kuznetsov, Ivan Kireev, and Andrey Savchenko. 2024.
\newblock {ESQA}: Event sequences question answering.
\newblock \emph{arXiv preprint arXiv:2407.12833}.

\bibitem[{Babaev et~al.(2022)Babaev, Ovsov, Kireev, Ivanova, Gusev, Nazarov, and Tuzhilin}]{babaev2022coles}
Dmitrii Babaev, Nikita Ovsov, Ivan Kireev, Maria Ivanova, Gleb Gusev, Ivan Nazarov, and Alexander Tuzhilin. 2022.
\newblock Coles: Contrastive learning for event sequences with self-supervision.
\newblock In \emph{Proceedings of the 2022 International Conference on Management of Data}, pages 1190--1199.

\bibitem[{Babaev et~al.(2019)Babaev, Savchenko, Tuzhilin, and Umerenkov}]{babaev2019rnn}
Dmitrii Babaev, Maxim Savchenko, Alexander Tuzhilin, and Dmitrii Umerenkov. 2019.
\newblock Et-rnn: Applying deep learning to credit loan applications.
\newblock In \emph{Proceedings of the 25th ACM SIGKDD international conference on knowledge discovery \& data mining}, pages 2183--2190.

\bibitem[{Bagheri et~al.(2023)Bagheri, Giachanou, Mosteiro, and Verberne}]{bagheri2023natural}
Ayoub Bagheri, Anastasia Giachanou, Pablo Mosteiro, and Suzan Verberne. 2023.
\newblock Natural language processing and text mining (turning unstructured data into structured).
\newblock In \emph{Clinical Applications of Artificial Intelligence in Real-World Data}, pages 69--93. Springer.

\bibitem[{Bao et~al.(2023)Bao, Zhang, Zhang, Wang, Feng, and He}]{bao2023tallrec}
Keqin Bao, Jizhi Zhang, Yang Zhang, Wenjie Wang, Fuli Feng, and Xiangnan He. 2023.
\newblock {TallRec}: An effective and efficient tuning framework to align large language model with recommendation.
\newblock In \emph{Proceedings of the 17th ACM Conference on Recommender Systems}, pages 1007--1014.

\bibitem[{BehnamGhader et~al.(2024)BehnamGhader, Adlakha, Mosbach, Bahdanau, Chapados, and Reddy}]{behnamghader2024llmvec}
Parishad BehnamGhader, Vaibhav Adlakha, Marius Mosbach, Dzmitry Bahdanau, Nicolas Chapados, and Siva Reddy. 2024.
\newblock {LLM}2vec: Large language models are secretly powerful text encoders.
\newblock In \emph{First Conference on Language Modeling}.

\bibitem[{Clark et~al.(2019)Clark, Luong, Le, and Manning}]{clarkelectra}
Kevin Clark, Minh-Thang Luong, Quoc~V Le, and Christopher~D Manning. 2019.
\newblock Electra: Pre-training text encoders as discriminators rather than generators.
\newblock In \emph{International Conference on Learning Representations}.

\bibitem[{Dai et~al.(2023)Dai, Liu, Dou, Wang, Liu, Long, and Wen}]{Dai2023}
Shitong Dai, Jiongnan Liu, Zhicheng Dou, Haonan Wang, Lin Liu, Bo~Long, and Ji-Rong Wen. 2023.
\newblock \href {https://doi.org/10.1145/3580305.3599287} {Contrastive learning for user sequence representation in personalized product search}.
\newblock In \emph{Proceedings of the 29th ACM SIGKDD Conference on Knowledge Discovery and Data Mining (KDD '23)}, pages 380--389. ACM.

\bibitem[{Devlin et~al.(2019)Devlin, Chang, Lee, and Toutanova}]{devlin2019bert}
Jacob Devlin, Ming-Wei Chang, Kenton Lee, and Kristina Toutanova. 2019.
\newblock Bert: Pre-training of deep bidirectional transformers for language understanding.
\newblock In \emph{Proceedings of the 2019 conference of the North American chapter of the association for computational linguistics: human language technologies, volume 1 (long and short papers)}, pages 4171--4186.

\bibitem[{Du et~al.(2022)Du, Qian, Liu, Ding, Qiu, Yang, and Tang}]{du2022glm}
Zhengxiao Du, Yujie Qian, Xiao Liu, Ming Ding, Jiezhong Qiu, Zhilin Yang, and Jie Tang. 2022.
\newblock Glm: General language model pretraining with autoregressive blank infilling.
\newblock In \emph{Proceedings of the 60th Annual Meeting of the Association for Computational Linguistics (Volume 1: Long Papers)}, pages 320--335.

\bibitem[{Gui et~al.(2024)Gui, Chen, Zhang, Cao, Sun, Luo, and Tao}]{gui2024survey}
Jie Gui, Tuo Chen, Jing Zhang, Qiong Cao, Zhenan Sun, Hao Luo, and Dacheng Tao. 2024.
\newblock A survey on self-supervised learning: Algorithms, applications, and future trends.
\newblock \emph{IEEE Transactions on Pattern Analysis and Machine Intelligence}.

\bibitem[{Guo et~al.(2020)Guo, Guo, Jin, Kaul, Gotz, and Cao}]{guo2020surveyvisualanalysisevent}
Yi~Guo, Shunan Guo, Zhuochen Jin, Smiti Kaul, David Gotz, and Nan Cao. 2020.
\newblock \href {https://arxiv.org/abs/2006.14291} {Survey on visual analysis of event sequence data}.
\newblock \emph{Preprint}, arXiv:2006.14291.

\bibitem[{Hegselmann et~al.(2023)Hegselmann, Buendia, Lang, Agrawal, Jiang, and Sontag}]{hegselmann2023tabllm}
Stefan Hegselmann, Alejandro Buendia, Hunter Lang, Monica Agrawal, Xiaoyi Jiang, and David Sontag. 2023.
\newblock {TabLLM}: Few-shot classification of tabular data with large language models.
\newblock In \emph{International Conference on Artificial Intelligence and Statistics}, pages 5549--5581. PMLR.

\bibitem[{Jiang et~al.(2023)Jiang, Chen, Zhao, Chen, Ping, Tran, Xu, Zeng, and Chilimbi}]{jiang2023understanding}
Qian Jiang, Changyou Chen, Han Zhao, Liqun Chen, Qing Ping, Son~Dinh Tran, Yi~Xu, Belinda Zeng, and Trishul Chilimbi. 2023.
\newblock Understanding and constructing latent modality structures in multi-modal representation learning.
\newblock In \emph{Proceedings of the IEEE/CVF Conference on Computer Vision and Pattern Recognition}, pages 7661--7671.

\bibitem[{Jin et~al.(2024)Jin, Wang, Ma, Chu, Zhang, Shi, Chen, Liang, Li, Pan et~al.}]{jintime}
Ming Jin, Shiyu Wang, Lintao Ma, Zhixuan Chu, James~Y Zhang, Xiaoming Shi, Pin-Yu Chen, Yuxuan Liang, Yuan-Fang Li, Shirui Pan, and 1 others. 2024.
\newblock {Time-LLM}: Time series forecasting by reprogramming large language models.
\newblock In \emph{The Twelfth International Conference on Learning Representations}.

\bibitem[{Karpukhin and Savchenko(2024)}]{karpukhin2024detpp}
Ivan Karpukhin and Andrey Savchenko. 2024.
\newblock {DeTPP}: Leveraging object detection for robust long-horizon event prediction.
\newblock \emph{arXiv preprint arXiv:2408.13131}.

\bibitem[{Ke et~al.(2017)Ke, Meng, Finley, Wang, Chen, Ma, Ye, and Liu}]{ke2017lightgbm}
Guolin Ke, Qi~Meng, Thomas Finley, Taifeng Wang, Wei Chen, Weidong Ma, Qiwei Ye, and Tie-Yan Liu. 2017.
\newblock Lightgbm: A highly efficient gradient boosting decision tree.
\newblock \emph{Advances in neural information processing systems}, 30.

\bibitem[{Kolosnjaji et~al.(2016)Kolosnjaji, Zarras, Webster, and Eckert}]{system_call_sequences}
Bojan Kolosnjaji, Apostolis Zarras, George Webster, and Claudia Eckert. 2016.
\newblock Deep learning for classification of malware system call sequences.
\newblock In \emph{AI 2016: Advances in Artificial Intelligence: 29th Australasian Joint Conference}, pages 137--149.

\bibitem[{Liu et~al.(2025)Liu, Dou, Nie, Chen, Tang, Xu, and Wen}]{liu2025enhancing}
Jiongnan Liu, Zhicheng Dou, Jian-Yun Nie, Zhenlin Chen, Guoyu Tang, Sulong Xu, and Ji-Rong Wen. 2025.
\newblock Enhancing sequential personalized product search with external out-of-sequence knowledge.
\newblock \emph{ACM Transactions on Information Systems}, 43(4):1--25.

\bibitem[{Liu et~al.(2023)Liu, Liu, Guo, Chen, Huang, Zhao, Tang, Luo, and Weng}]{Liu2023XES3G5M}
Zitao Liu, Qiongqiong Liu, Teng Guo, Jiahao Chen, Shuyan Huang, Xiangyu Zhao, Jiliang Tang, Weiqi Luo, and Jian Weng. 2023.
\newblock \href {https://datasets-benchmarks-proceedings.neurips.cc/paper/2023/file/67fc628f17c2ad53621fb961c6bafcaf-Paper.pdf} {{XES3G5M}: A knowledge tracing benchmark dataset with auxiliary information}.
\newblock In \emph{Advances in Neural Information Processing Systems 36: Datasets and Benchmarks Track (NeurIPS 2023)}.

\bibitem[{Luetto et~al.(2025)Luetto, Garuti, Sangineto, Forni, and Cucchiara}]{luetto2025one}
Simone Luetto, Fabrizio Garuti, Enver Sangineto, Lorenzo Forni, and Rita Cucchiara. 2025.
\newblock One transformer for all time series: Representing and training with time-dependent heterogeneous tabular data.
\newblock \emph{Machine Learning}, 114(6):1--27.

\bibitem[{Mollaev et~al.(2025)Mollaev, Kostin, Postnova, Karpukhin, Kireev, Gusev, and Savchenko}]{mollaev2025multimodalbankingdatasetunderstanding}
Dzhambulat Mollaev, Alexander Kostin, Maria Postnova, Ivan Karpukhin, Ivan Kireev, Gleb Gusev, and Andrey Savchenko. 2025.
\newblock \href {https://arxiv.org/abs/2409.17587} {Multimodal banking dataset: Understanding client needs through event sequences}.
\newblock \emph{Preprint}, arXiv:2409.17587.

\bibitem[{Muennighoff et~al.(2024)Muennighoff, Hongjin, Wang, Yang, Wei, Yu, Singh, and Kiela}]{muennighoff2024generative}
Niklas Muennighoff, SU~Hongjin, Liang Wang, Nan Yang, Furu Wei, Tao Yu, Amanpreet Singh, and Douwe Kiela. 2024.
\newblock Generative representational instruction tuning.
\newblock In \emph{ICLR 2024 Workshop: How Far Are We From AGI}.

\bibitem[{Muennighoff et~al.(2023)Muennighoff, Tazi, Magne, Schwenk, Lample, Douze, Aizawa, and Grave}]{muennighoff2023mteb}
Niklas Muennighoff, Nouamane Tazi, Lo{\"\i}c Magne, Holger Schwenk, Guillaume Lample, Matthijs Douze, Akiko Aizawa, and Edouard Grave. 2023.
\newblock {MTEB: Massive Text Embedding Benchmark}.
\newblock In \emph{Proceedings of the 17th Conference of the European Chapter of the Association for Computational Linguistics (EACL)}.

\bibitem[{Oord et~al.(2018)Oord, Li, and Vinyals}]{oord2018representation}
Aaron van~den Oord, Yazhe Li, and Oriol Vinyals. 2018.
\newblock Representation learning with contrastive predictive coding.
\newblock \emph{arXiv preprint arXiv:1807.03748}.

\bibitem[{Osin et~al.(2025)Osin, Udovichenko, Moskvoretskii, Shvetsov, and Burnaev}]{osin2025ebeseasybenchmarkingevent}
Dmitry Osin, Igor Udovichenko, Viktor Moskvoretskii, Egor Shvetsov, and Evgeny Burnaev. 2025.
\newblock \href {https://arxiv.org/abs/2410.03399} {{EBES}: Easy benchmarking for event sequences}.
\newblock \emph{Preprint}, arXiv:2410.03399.

\bibitem[{Radford et~al.(2021)Radford, Kim, Hallacy, Ramesh, Goh, Agarwal, Sastry, Askell, Mishkin, Clark et~al.}]{radford2021learning}
Alec Radford, Jong~Wook Kim, Chris Hallacy, Aditya Ramesh, Gabriel Goh, Sandhini Agarwal, Girish Sastry, Amanda Askell, Pamela Mishkin, Jack Clark, and 1 others. 2021.
\newblock Learning transferable visual models from natural language supervision.
\newblock In \emph{International conference on machine learning}, pages 8748--8763. PmLR.

\bibitem[{Rosbank(2021)}]{rosbank2021churn}
Rosbank. 2021.
\newblock Churn prediction challenge.
\newblock \url{https://boosters.pro/championship/rosbank1/overview}.
\newblock Accessed: 2025-07-04.

\bibitem[{Ruan et~al.(2024)Ruan, Lan, Ma, Dong, He, and Feng}]{ruan2024languagemodelingtabulardata}
Yucheng Ruan, Xiang Lan, Jingying Ma, Yizhi Dong, Kai He, and Mengling Feng. 2024.
\newblock \href {https://arxiv.org/abs/2408.10548} {Language modeling on tabular data: A survey of foundations, techniques and evolution}.
\newblock \emph{Preprint}, arXiv:2408.10548.

\bibitem[{Sberbank(2021{\natexlab{a}})}]{sberbank2021gender}
Sberbank. 2021{\natexlab{a}}.
\newblock Python and analyze data: Final project (gender).
\newblock \url{https://www.kaggle.com/competitions/python-and-analyze-data-final-project}.
\newblock Accessed: 2025-07-04.

\bibitem[{Sberbank(2021{\natexlab{b}})}]{sberbank2021age}
Sberbank. 2021{\natexlab{b}}.
\newblock Sirius age group competition.
\newblock \url{https://ods.ai/competitions/sberbank-sirius-lesson}.
\newblock Accessed: 2025-07-04.

\bibitem[{Shchur et~al.(2020)Shchur, Bilo{\v{s}}, and G{\"u}nnemann}]{shchurintensity}
Oleksandr Shchur, Marin Bilo{\v{s}}, and Stephan G{\"u}nnemann. 2020.
\newblock Intensity-free learning of temporal point processes.
\newblock In \emph{International Conference on Learning Representations}.

\bibitem[{Shestov et~al.(2025)Shestov, Zoloev, Makarenko, Orlov, Fadeev, Kireev, and Savchenko}]{shestov2025llm4es}
Aleksei Shestov, Omar Zoloev, Maksim Makarenko, Mikhail Orlov, Egor Fadeev, Ivan Kireev, and Andrey Savchenko. 2025.
\newblock {LLM4ES}: Learning user embeddings from event sequences via large language models.
\newblock \emph{arXiv preprint arXiv:2508.05688}.

\bibitem[{Shi et~al.(2023)Shi, Xue, Wang, Zhou, Zhang, Zhou, and Mei}]{shi2023language}
Xin Shi, Shizhe Xue, Kun Wang, Feng Zhou, Jiawei Zhang, Jingren Zhou, and Hongyu Mei. 2023.
\newblock {Language Models Can Improve Event Prediction by Few-Shot Abductive Reasoning}.
\newblock In \emph{Advances in Neural Information Processing Systems (NeurIPS)}, volume~36, pages 29532--29557.

\bibitem[{Shou et~al.(2023)Shou, Bhattacharjya, Gao, Subramanian, Hassanzadeh, and Bennett}]{shou2023pairwise}
Xiaohan Shou, Debarun Bhattacharjya, Tong Gao, Devika Subramanian, Oktie Hassanzadeh, and Kristin~P. Bennett. 2023.
\newblock {Pairwise Causality Guided Transformers for Event Sequences}.
\newblock In \emph{Advances in Neural Information Processing Systems (NeurIPS)}, volume~36, pages 46520--46533.

\bibitem[{Skalski et~al.(2023)Skalski, Sutton, Burrell, Perez, and Wong}]{skalski2023towards}
Piotr Skalski, David Sutton, Stuart Burrell, Iker Perez, and Jason Wong. 2023.
\newblock Towards a foundation purchasing model: Pretrained generative autoregression on transaction sequences.
\newblock In \emph{Proceedings of the Fourth ACM International Conference on AI in Finance}, pages 141--149.

\bibitem[{Sun et~al.(2024)Sun, Li, Li, and Hong}]{suntest}
Chenxi Sun, Hongyan Li, Yaliang Li, and Shenda Hong. 2024.
\newblock Test: Text prototype aligned embedding to activate llm's ability for time series.
\newblock In \emph{The Twelfth International Conference on Learning Representations}.

\bibitem[{Team et~al.(2025)Team, Kamath, Ferret, Pathak, Vieillard, Merhej, Perrin, Matejovicova, Ram{\'e}, Rivi{\`e}re et~al.}]{team2025gemma}
Gemma Team, Aishwarya Kamath, Johan Ferret, Shreya Pathak, Nino Vieillard, Ramona Merhej, Sarah Perrin, Tatiana Matejovicova, Alexandre Ram{\'e}, Morgane Rivi{\`e}re, and 1 others. 2025.
\newblock Gemma 3 technical report.
\newblock \emph{arXiv preprint arXiv:2503.19786}.

\bibitem[{Touvron et~al.(2023)Touvron, Lavril, Izacard, Martinet, Lachaux, Lacroix, Rozi{\`e}re, Goyal, Hambro, Azhar et~al.}]{touvron2023llama}
Hugo Touvron, Thibaut Lavril, Gautier Izacard, Xavier Martinet, Marie-Anne Lachaux, Timoth{\'e}e Lacroix, Baptiste Rozi{\`e}re, Naman Goyal, Eric Hambro, Faisal Azhar, and 1 others. 2023.
\newblock Llama: Open and efficient foundation language models.
\newblock \emph{arXiv preprint arXiv:2302.13971}.

\bibitem[{Touvron et~al.(2024)}]{touvron2024llama3}
Hugo Touvron and 1 others. 2024.
\newblock \href {https://arxiv.org/abs/2407.21783} {The {LLaMA} 3 herd of models}.
\newblock \emph{arXiv preprint arXiv:2407.21783}.

\bibitem[{Udovichenko et~al.(2024)Udovichenko, Shvetsov, Divitsky, Osin, Trofimov, Sukharev, Glushenko, Berestnev, and Burnaev}]{Udovichenko_2024}
Igor Udovichenko, Egor Shvetsov, Denis Divitsky, Dmitry Osin, Ilya Trofimov, Ivan Sukharev, Anatoliy Glushenko, Dmitry Berestnev, and Evgeny Burnaev. 2024.
\newblock \href {https://doi.org/10.1109/access.2024.3349497} {Seqnas: Neural architecture search for event sequence classification}.
\newblock \emph{IEEE Access}, 12:3898–3909.

\bibitem[{Wang et~al.(2024{\natexlab{a}})Wang, Yang, Huang, Yang, Majumder, and Wei}]{wang2024multilingual}
Liang Wang, Nan Yang, Xiaolong Huang, Linjun Yang, Rangan Majumder, and Furu Wei. 2024{\natexlab{a}}.
\newblock Multilingual e5 text embeddings: A technical report.
\newblock \emph{arXiv preprint arXiv:2402.05672}.

\bibitem[{Wang et~al.(2024{\natexlab{b}})Wang, Fu, Xu, Ma, Xu, Du, Lu, Gao, Wu, and Chen}]{wang2024twin}
Yue Wang, Tianfan Fu, Yinlong Xu, Zihan Ma, Hongxia Xu, Bang Du, Yingzhou Lu, Honghao Gao, Jian Wu, and Jintai Chen. 2024{\natexlab{b}}.
\newblock Twin-gpt: digital twins for clinical trials via large language model.
\newblock \emph{ACM Transactions on Multimedia Computing, Communications and Applications}.

\bibitem[{Weiss and Hirsh(1998)}]{Rare_Events}
Gary~M. Weiss and Haym Hirsh. 1998.
\newblock Learning to predict rare events in event sequences.
\newblock In \emph{KDD}, pages 359--363.

\bibitem[{Yang et~al.(2025)Yang, Li, Yang, Zhang, Hui, Zheng, Yu, Gao, Huang, Lv et~al.}]{yang2025qwen3}
An~Yang, Anfeng Li, Baosong Yang, Beichen Zhang, Binyuan Hui, Bo~Zheng, Bowen Yu, Chang Gao, Chengen Huang, Chenxu Lv, and 1 others. 2025.
\newblock Qwen3 technical report.
\newblock \emph{arXiv preprint arXiv:2505.09388}.

\bibitem[{Yeshchenko and Mendling(2022)}]{yeshchenko2022surveyapproacheseventsequence}
Anton Yeshchenko and Jan Mendling. 2022.
\newblock \href {https://arxiv.org/abs/2202.07941} {A survey of approaches for event sequence analysis and visualization using the esevis framework}.
\newblock \emph{Preprint}, arXiv:2202.07941.

\bibitem[{Yin et~al.(2023)Yin, Xie, Qin, Ding, Feng, Li, and Lin}]{yin2023heterogeneous}
Bin Yin, Junjie Xie, Yu~Qin, Zixiang Ding, Zhichao Feng, Xiang Li, and Wei Lin. 2023.
\newblock Heterogeneous knowledge fusion: A novel approach for personalized recommendation via {LLM}.
\newblock In \emph{Proceedings of the 17th ACM Conference on Recommender Systems}, pages 599--601.

\bibitem[{Yu et~al.(2025)Yu, Guo, Fu, and Jin}]{yu2025eventformer}
Shuai Yu, Dongjie Guo, Yanjie Fu, and Wen Jin. 2025.
\newblock {EventFormer: A Hierarchical Neural Point Process Framework for Spatio-Temporal Clustering Events Prediction}.
\newblock \emph{Journal of Big Data}, 12(1):162.

\bibitem[{Zhang et~al.(2023)Zhang, Liu, Wang, Li, Chen, and Wang}]{zhang2023fatatrans}
Wei Zhang, Chao Liu, Yuxuan Wang, Tao Li, Junjie Chen, and Weiqiang Wang. 2023.
\newblock {FATA-Trans: Field and Time-Aware Transformer for Sequential Tabular Data}.
\newblock In \emph{Proceedings of the 32nd ACM International Conference on Information and Knowledge Management (CIKM)}.

\bibitem[{Zhang et~al.(2025)Zhang, Li, Long, Zhang, Lin, Yang, Xie, Yang, Liu, Lin et~al.}]{zhang2025qwen3}
Yanzhao Zhang, Mingxin Li, Dingkun Long, Xin Zhang, Huan Lin, Baosong Yang, Pengjun Xie, An~Yang, Dayiheng Liu, Junyang Lin, and 1 others. 2025.
\newblock Qwen3 embedding: Advancing text embedding and reranking through foundation models.
\newblock \emph{arXiv preprint arXiv:2506.05176}.

\bibitem[{Zheng et~al.(2024)Zheng, Chao, Qiu, Zhu, and Xiong}]{zheng2024harnessing}
Zhi Zheng, Wen-Shuo Chao, Zhaopeng Qiu, Hengshu Zhu, and Hui Xiong. 2024.
\newblock Harnessing large language models for text-rich sequential recommendation.
\newblock In \emph{The Web Conference 2024}.

\end{thebibliography}

\clearpage

\appendix

\section{Experimental Details}
\label{sec:appendixset}
\textbf{Experimental Setup.} For all experiments, natural language descriptions were generated using several instruction-tuned large language models of different types and scales, including \texttt{Gemma-3-27B-Instruct}, \texttt{Gemma-2-27B-Instruct}~\citep{team2025gemma}, and \texttt{Qwen3-Instruct-32B} and \texttt{Qwen3-Instruct-4B}~\citep{zhang2025qwen3}. To encode the resulting behavioral descriptions into fixed-length vectors, we employed the \texttt{Qwen3-Embedding-8B}~\citep{zhang2025qwen3} model, a multilingual encoder optimized for semantic retrieval. The transaction encoder was instantiated as a GRU-based model trained under the CoLES self-supervised framework, serving as the base sequence encoder across all contrastive alignment heads. Both training and inference for the full pipeline—including LLM prompting, embedding computation, and contrastive alignment—were performed on eight NVIDIA Tesla A100 80 GB GPUs.\\
\textbf{Pipeline Details.} A lightweight GRU-based transaction encoder  was pretrained under the CoLES objective and later tuned via lightweight contrastive alignment heads, while keeping the text encoders frozen. Alignment was performed between sequence embeddings and combined textual–statistical embeddings derived from LLM-generated descriptions enriched with numerical features. For fair comparison across downstream models, feature selection was applied to all representations on the validation set, fixing the embedding dimensionality to 512 or 1024.

\section{Additional Experiments}
\label{sec:appendixexp}

\paragraph{Impact of Text Embedding Extraction Strategy.}

\begin{table}[h]
\centering
\resizebox{0.95\columnwidth}{!}{
\begin{tabular}{llccc}
\hline
\textbf{Contrastive Head} & \textbf{Method} & \textbf{Churn (AUC)} & \textbf{Age Group (Acc)} & \textbf{Gender (AUC)} \\
\hline
\methodname[\ref{eq:softmax}] & LLM encoder & 0.870 ± 0.005 & 0.665 ± 0.004 & 0.895 ± 0.002 \\
\methodname[\ref{eq:softmax}] & MeanPool & 0.871 ± 0.003 & 0.663 ± 0.003 & 0.896 ± 0.001 \\
\hline
\end{tabular}
}
\caption{Ablation: Comparing description embeddings obtained from a dedicated sentence encoder (ours) vs.\ directly from the generator LLM.}
\label{tab:ablation_embedding_source}
\end{table}

We compare two strategies for obtaining text embeddings from behavioral descriptions. In our default setup, we use a frozen sentence encoder (Qwen3-Embedding-8B) to compute embeddings, separating the generation and encoding processes. As an alternative, we extract the embedding directly from the generator LLM (Gemma-3-27B) by mean-pooling hidden states. Following recent work suggesting that mean pooling over all token embeddings outperforms using the EOS token~\citep{behnamghader2024llmvec, muennighoff2024generative}, we average hidden states over the final $k=8$ transformer layers. Both versions provides similar embeddings quality.

\paragraph{Language Model Choice for Description Generation.}

\begin{table}[h]
\centering
\resizebox{\columnwidth}{!}{
\begin{tabular}{lccc}
\hline
\textbf{Metric} 
& \textbf{Gemma 3 4B} 
& \textbf{Qwen 3 32B} 
& \textbf{Gemma 3 27B} \\ 
\hline
Churn (AUC) & 0.872 ± 0.003 & 0.870 ± 0.005 & 0.869 ± 0.005 \\
Age Group (Acc) & 0.658 ± 0.007 & 0.665 ± 0.005 & 0.657 ± 0.004 \\
Gender (AUC) & 0.897 ± 0.005 & 0.895 ± 0.002 & 0.898 ± 0.003 \\
\hline
\end{tabular}
}
\caption{Ablation: Effect of language model choice for generating behavioral descriptions.}
\label{tab:ablation_llm_choice}
\end{table}

Table~\ref{tab:ablation_llm_choice} examines how the choice of large language model for generating behavioral descriptions influences downstream performance. Among the compared generators, Qwen 3 32B achieves the highest average accuracy across tasks, leading on age group classification (0.665 Acc) and delivering competitive results on churn and gender prediction. Gemma 3 27B attains the best gender AUC (0.898) and remains comparable in other metrics, confirming that mid-scale models with strong instruction tuning can match much larger ones. In contrast, the compact Gemma 3 4B variant underperforms across all tasks. Overall, the results suggest that richer generative capacity and stronger instruction following in the LLM used for description synthesis translate into more informative and transferable sequence representations

\paragraph{Text Embedding Model Selection.}

\begin{table}[h]
\centering
\resizebox{\columnwidth}{!}{
\begin{tabular}{llccc}
\hline
\textbf{Contrastive Head} & \textbf{Text Encoder} & \textbf{Churn (AUC)} & \textbf{Age Group (Acc)} & \textbf{Gender (AUC)} \\
\hline
\methodname[\ref{eq:softmax}] & mE5-large-instr & 0.869 ± 0.005 & 0.662 ± 0.008 & 0.896 ± 0.002 \\
\methodname[\ref{eq:softmax}] & Qwen3-Emb-0.6B  & 0.862 ± 0.005 & 0.662 ± 0.004 & 0.899 ± 0.003 \\
\methodname[\ref{eq:softmax}] & Qwen3-Emb-8B    & 0.870 ± 0.005 & 0.665 ± 0.005 & 0.895 ± 0.002 \\
\hline
\end{tabular}
}
\caption{Ablation: Impact of the text embedding model on downstream task performance.}
\label{tab:ablation_text_encoder}
\end{table}

Table~\ref{tab:ablation_text_encoder} analyzes how the choice of frozen text encoder for obtaining $z_i^{\text{text}}$ affects downstream task performance. We compare mE5-large-instruct~\citep{wang2024multilingual} with two Qwen3 embedding variants~\citep{yang2025qwen3}: a compact Qwen3-Emb-0.6B and a larger Qwen3-Emb-8B. The results show that all encoders perform comparably, with differences within a narrow margin of 0.5–1.0 pp across tasks. mE5-large-instruct yields the highest AUC on churn prediction (0.869), while Qwen3-Emb-8B slightly leads in age classification (0.665 Acc). The Qwen3-Emb-0.6B model achieves the best gender AUC (0.899), despite being the smallest. 
\paragraph{Runtime Analysis}

\begin{figure}[h]
    \centering
    \includegraphics[width=\linewidth]{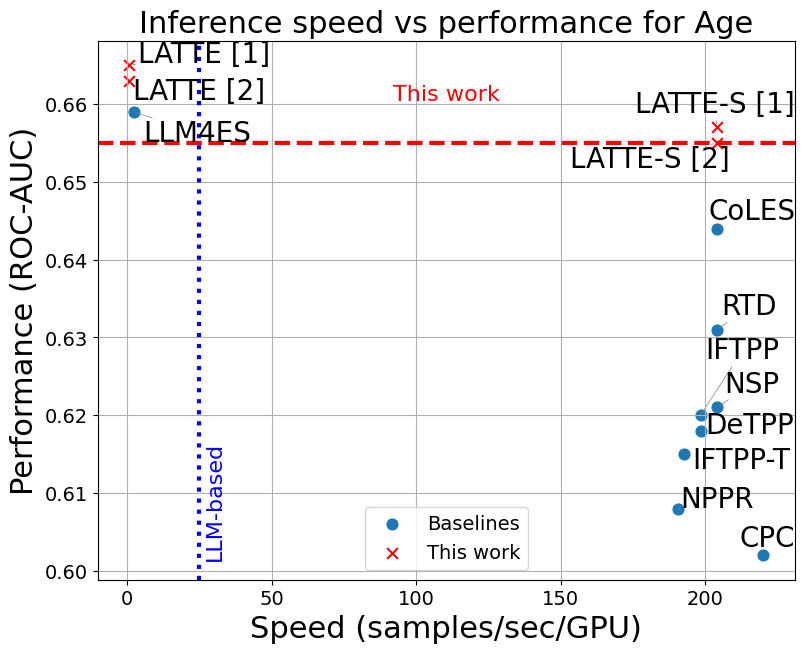} 
    \caption{
        Figure of Merit comparison of model performance in ROC-AUC and inference speed in samples per second per GPU on Age dataset.
    }
    \label{fig:age}
\end{figure}

Figure~\ref{fig:age} presents results for the Age prediction task. We observe trends consistent with those in the Churn dataset (Figure~\ref{fig:churn}): generative models attain the highest accuracy (up to 0.665) but exhibit limited inference throughput. In contrast, lightweight LATTE-S variants achieve a favorable trade-off, maintaining competitive performance (0.57 ROC-AUC) while delivering significantly higher inference speeds, exceeding 200 samples per second.

\section{Text Generation Protocol}
\label{sec:appendix_generation}

We generate natural language descriptions from statistical user summaries by prompting an instruction-tuned LLM. Table~\ref{tab:llm_prompt} illustrates the full prompt used for behavioral description generation, while Table~\ref{tab:llm_example} provides an example output produced by the model.

\begin{table*}[t]
\centering
\begin{tabular}{p{\textwidth}}
\hline
\textbf{System:} \textit{You are an expert in financial transaction analysis. Your task is to generate clear, structured, and concise descriptions of user financial behavior based on given transaction data. Use data-driven insights and avoid speculation.} \\
\\
\textbf{Guidelines:}
\begin{itemize}
\item Do \textbf{not} include phrases like “Here’s a financial behavior description for User X”
\item Start \textbf{directly} with behavioral insights
\item Interpret numbers into \textbf{patterns} (e.g., burst spending, routine payments)
\item Highlight \textbf{spending habits, risk factors, financial consistency}
\item Avoid unsupported assumptions; maintain clarity and conciseness
\end{itemize}
\\
\textbf{User:} Below is a summary of a user’s transaction history:
\begin{verbatim}
- Number of transactions: 226
- Active transaction period: 444 days
- Avg transactions per day: 0.51
- Avg transaction interval: 1.97 days
- Top MCCs: Supermarkets, cash withdrawals, telecommunications
- Share of transactions in Top 1 MCC: 0.04
- Avg amount for Top 1 MCC: 2770.47 RUB
- Avg amount for Top 2 MCC: 45969.41 RUB
- Avg amount for Top 3 MCC: 8139.52 RUB
- Share of days with transactions: 0.36
- Total income: 2,706,553.10 RUB
- Total expenses: 3,956,007.97 RUB
- Avg outgoing amount: 17,981.85 RUB
- Avg incoming amount: 451,092.18 RUB
\end{verbatim}
\textbf{Instructions:}
\begin{itemize}
\item Analyze behavioral patterns
\item Identify transaction regularity and category reliance
\item Assess potential risk factors and financial planning traits
\item Write in a structured and engaging way while staying factual
\end{itemize}
\\
\hline
\end{tabular}
\caption{Prompt template for behavioral description generation}
\label{tab:llm_prompt}
\end{table*}

\begin{table*}[t]
\centering
\begin{tabular}{p{\textwidth}}
\hline
\textbf{Output:} \\
This user demonstrates a \textbf{low-frequency, high-variance spending pattern} over a substantial period (444 days). Transactions occur on approximately 36\% of days, averaging one every two days, indicating infrequent but potentially substantial financial activity.

Spending is heavily weighted towards \textbf{cash withdrawals from financial institutions} (4\% of transactions, average 45969.41 RUB), suggesting a preference for cash-based expenses or large withdrawals for specific purposes. Significant spending also occurs at \textbf{grocery stores/supermarkets} (average 8139.52 RUB), representing a consistent, though smaller, portion of overall expenditure. A smaller number of transactions are related to \textbf{phone services} (average 2770.47 RUB).

\textbf{Expenses significantly exceed income} (3,956,007.97 RUB vs. 2,706,553.10 RUB), implying reliance on savings, credit, or other income sources not captured in this data. The high average outgoing transaction amount (17,981.85 RUB) coupled with a large average incoming amount (451,092.18 RUB) suggests \textbf{irregular but substantial income events} are occurring, potentially offsetting consistent overspending.

The combination of infrequent transactions, high cash withdrawals, and a substantial expense-to-income imbalance points to \textbf{potentially inconsistent financial management}. While not indicative of reckless spending, the reliance on infrequent income and high cash usage warrants further investigation to determine financial stability and potential risk factors. There’s a lack of diversification in spending, with the top three MCCs dominating activity.
\\
\hline
\end{tabular}
\caption{Example output generated from the LLM}
\label{tab:llm_example}
\end{table*}

\end{document}